\documentclass[11pt,a4paper]{article}
\usepackage{times,latexsym}
\usepackage{url}
\usepackage[T1]{fontenc}
\usepackage{multirow}
\usepackage{bm}
\usepackage{microtype}
 \usepackage{amsmath}
\usepackage{graphicx}
\usepackage{booktabs}
\usepackage{arydshln}
\usepackage{CJKutf8}
\usepackage{amssymb}
\usepackage{paralist}

\newcommand{\paratitle}[1]{\vspace{1.5ex}\noindent \textbf{#1}}
\usepackage[acceptedWithA]{tacl2021v1}
\usepackage{xspace,mfirstuc,tabulary}

\newif\iftaclinstructions
\taclinstructionsfalse 
\iftaclinstructions
\renewcommand{\confidential}{}
\renewcommand{\anonsubtext}{(No author info supplied here, for consistency with
TACL-submission anonymization requirements)}
\newcommand{\instr}
\fi

\iftaclpubformat 

\else

\fi

\title{Modeling Unified Semantic Discourse Structure\\ for High-quality Headline Generation}

\author{
Minghui Xu$^{1}$, Hao Fei$^{2}$, Fei Li$^{1}$, Shengqiong Wu$^{2}$, Rui Sun$^{3}$, Chong Teng$^{1}$, Donghong Ji$^{1}$
\\
$^{1}$ School of Cyber Science and Engineering, Wuhan University
\\
$^{2}$ National University of Singapore \quad $^{3}$ Leshan Normal University\\
\texttt{\{frank.xu, ruisun, tengchong, dhji\}@whu.edu.cn}\\
\texttt{haofei37@nus.edu.sg, foxlf823@gmail.com, swu@u.nus.edu}
}

\date{}

\begin{document}
\begin{CJK}{UTF8}{gbsn}
\maketitle
\begin{abstract}

Headline generation aims to summarize a long document with a short, catchy title that reflects the main idea.
This requires accurately capturing the core document semantics, which is challenging due to the lengthy and background information-rich nature of the texts.
In this work, We propose using a unified semantic discourse structure (S$^3$) to represent document semantics, achieved by combining document-level rhetorical structure theory (RST) trees with sentence-level abstract meaning representation (AMR) graphs to construct S$^3$ graphs. The hierarchical composition of sentence, clause, and word intrinsically characterizes the semantic meaning of the overall document.
We then develop a headline generation framework, in which the S$^3$ graphs are encoded as contextual features.
To consolidate the efficacy of S$^3$ graphs, we further devise a hierarchical structure pruning mechanism to dynamically screen the redundant and nonessential nodes within the graph.
Experimental results on two headline generation datasets demonstrate that our method outperforms existing state-of-art methods consistently.
Our work can be instructive for a broad range of document modeling tasks, more than headline or summarization generation.
\end{abstract}

\section{Introduction}
The task of headline generation \cite{OverDH07,cohn-lapata-2008-sentence} has been proposed with the aim of automatically generating a concise and catchy sentence that summarizes the main idea or topic of a document.
In the last decade, the research of headline generation has received substantial attention \cite{sun-etal-2015-event,shen2017recent,matsumaru-etal-2020-improving}.
Early headline generation is usually regarded as a variant of document summarization tasks, solved with various methods \cite{cohn-lapata-2008-sentence,sun-etal-2015-event,shen2017recent}.
Subsequent studies \cite{xu-etal-2019-clickbait,song2020attractive} have shown that a headline does not need to include all the points of a document as in the summarization task.

Later on, the research of headline generation shifts attention to \emph{truthfulness} and \emph{attractiveness}.
In particular, the former emphasizes the accuracy and honesty of the headline in representing the raw content, while the latter requires the headlines to catch the reader's attention and interest  \cite{matsumaru-etal-2020-improving, song2020attractive,wang-etal-2022-timestep}. 
Nevertheless, all the existing explorations overlook the intrinsic characteristics of the document, which inevitably hinders the improvement of task performance.

\begin{figure}[!t]
\centering
\includegraphics[width=0.98\columnwidth]{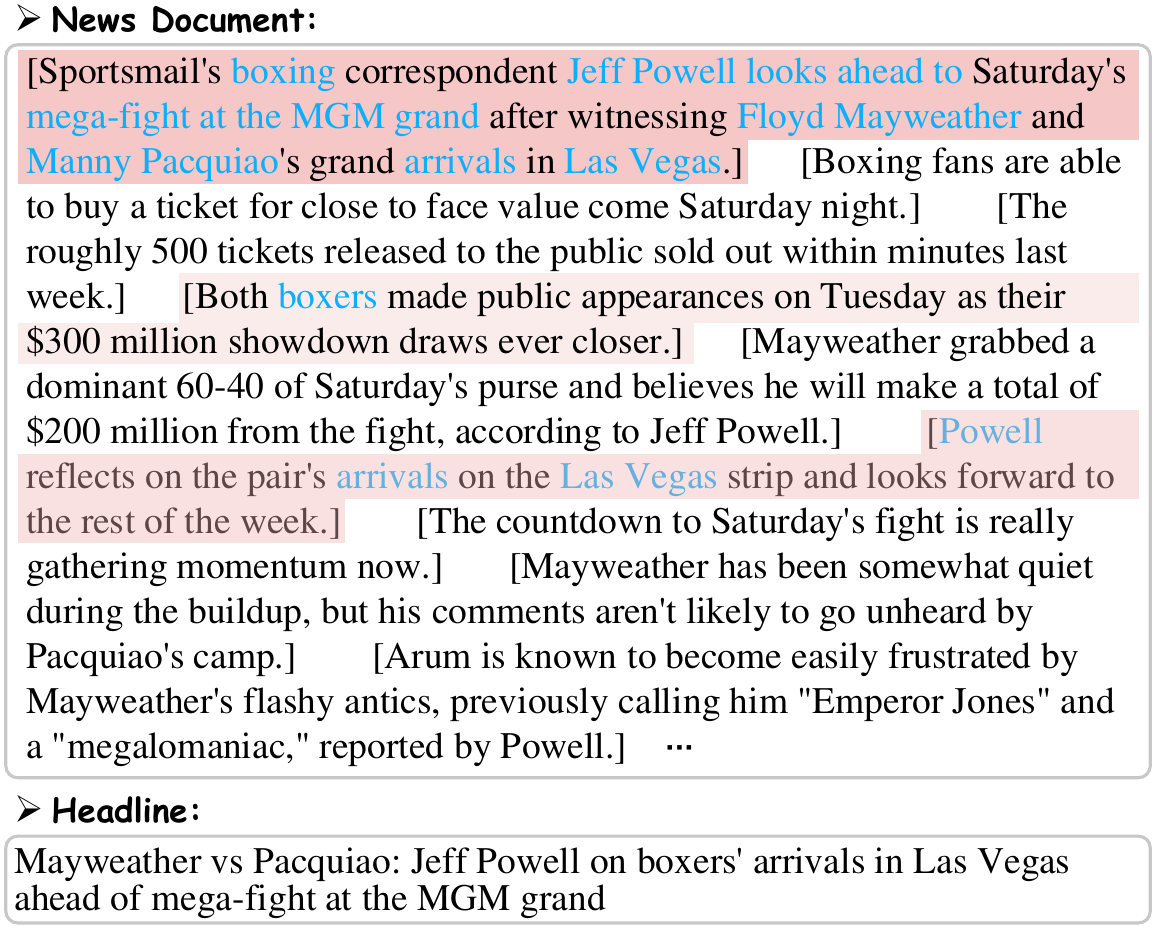}
\caption{
An example of news and corresponding headline. The news is segmented into sentences.
We mark the decisive sentences in red color, where the keywords used in the headline are highlighted in blue color.
}
\label{intro}
\vspace{-3mm}
\end{figure}

Document texts consist of a considerable number of subordinate sentences or clauses, thus containing lengthy and mixed information.
Consequently, the key clues that describe the headline are scattered around the whole document broadly, as illustrated in Figure \ref{intro}.
Actually, documents come with hierarchical structures in two granularities, i.e., document level and sentence level.
Document texts are essentially organized with the underlying discourse structures, where each sentence piece serves a different role in the document \cite{gerani-etal-2014-abstractive, Lin2019HierarchicalPN}.
Unfortunately, existing methods directly encode the document texts in the traditional sequential manner, while failing to consider the nature of discourse structure.
In addition, the sentence is the atomic information unit within a document, which makes it important to coordinate the sentence-level local features with the overall global contexts.

\begin{figure*}[!t]
\centering
\includegraphics[width=0.98\textwidth]{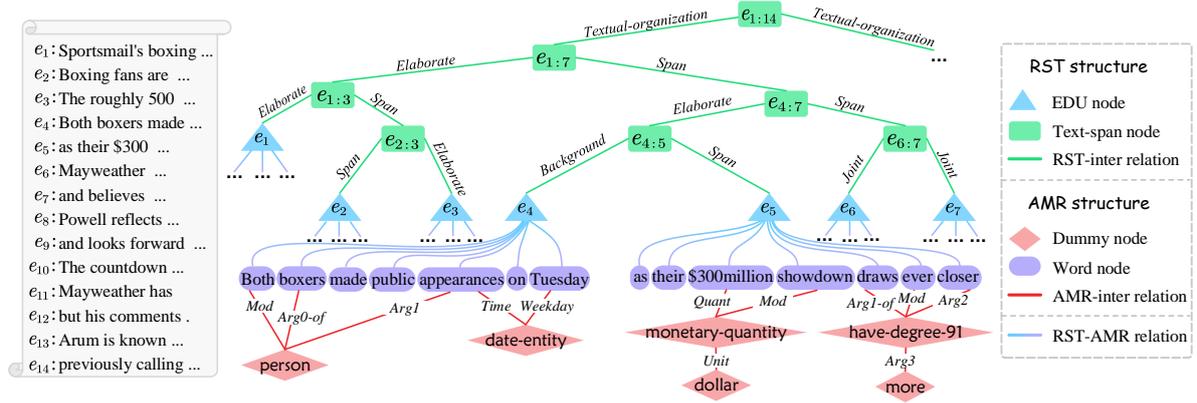}
\caption{
Illustration of our proposed unified semantic discourse structure (S$^3$) of the corresponding document (left), which is composed of the document-level RST structure and sentence-level AMR structure.
In the RST structure, the non-leaf text-span nodes are virtual nodes, while the terminal EDU nodes are sentences or clauses.
In the AMR structure, dummy nodes represent the virtual concepts, and word nodes connect to the corresponding EDU node via the RST-AMR relation.
}
\label{discourse-graph}
\vspace{-3mm}
\end{figure*}

Based on the above observation, we consider modeling the semantic discourse structure information to achieve high-quality headline generation.
We represent the input documents with an integrated hierarchical structure, which is composed of the document-level rhetorical structure theory (RST) \citep{mann1988rhetorical} and the sentence-level abstract meaning representation (AMR) \citep{banarescu2013abstract}, as demonstrated in Figure \ref{discourse-graph}.
RST depicts how sub-sentences and clauses (aka, elementary discourse unit, EDU) are organized into the discourse, 
and provides intrinsic features indicating which parts of the document are important.
In contrast, AMR is a graph formalism representing the core semantic meaning of sentence-level texts.

By joining RST and AMR structures on EDUs, we construct a unified \underline{s}emantic di\underline{s}course \underline{s}tructure (namely S$^3$).
Besides S$^3$, we further implement a headline generation framework that models the S$^3$ graph features.
As shown in Figure \ref{fig:framework}, the system takes an encoder-decoder paradigm.
During encoding, a pre-trained language model (PLM) first learns the contextual representations for the input document.
Then, we use the graph attention network (GAT) \citep{velivckovic2017graph} to perform information propagation over the S$^3$ graph.


Moreover, considering that the majority of information in the document graph is not beneficial to the headline generation, we thus consider a dynamic structure pruning over S$^3$, filtering those redundant and task-irrelevant nodes based on the reinforcement learning (RL) \cite{williams1992simple} and
each decision is made under the guidance of end-task supervision, resulting in a compact yet highly-effective discourse structure.


We carry out experiments on two benchmark datasets, including CNNDM-DH \cite{Nallapati2016abstractive} and DM-DHC \cite{song2020attractive}.
Results demonstrate that our method outperforms current state-of-the-art (SoTA) models by significant margins.
Ablation studies show the importance to integrate the proposed semantic discourse structure features for high-quality headline generation.
Further analyses reveal that the discourse structure information especially helps in handling long documents, and also the proposed dynamic pruning mechanism effectively enhances the efficacy of the discourse structure.
Overall, this paper contributes by three folds:
\begin{compactitem}

\item[$\bullet$] We for the first time construct a type of unified semantic discourse structure for representing the core semantics of documents for high-quality headline generation.

\item[$\bullet$] We enhance the efficacy of the semantic discourse structure representation by proposing a novel hierarchical graph pruning mechanism based on the reinforcement learning technique.

\item[$\bullet$] Our system achieves the SoTA performances on two headline generation benchmark datasets.
Our code will be publicly available to facilitate related research.

\end{compactitem}


\section{Related work}

\subsection{Headline Generation} 
Automatic headline generation, a.k.a., title generation, is pioneered by \citet{kennedy2000automatic} and \citet{jin-hauptmann-2001-automatic}.
As an important task in the natural language generation track, the task has received much research attention.
In prior work, headline generation has been extensively modeled as an article (or abstract) summarization problem.
Following the text summarization line, the methods are classified into extractive and abstractive paradigms.
Extractive summarization methods produce headlines texts by extracting key sentences within the documents and ranking them according to the topic relevance \cite{cao2015ranking,cheng-lapata-2016-neural,dong-etal-2018-banditsum,gu-etal-2022-memsum}.
In contrast, abstractive-based methods directly generate texts by understanding the article and summarizing the idea with new expressions.
Combining with the neural sequence-to-sequence (seq2seq) models \citep{bahdanau2014neural,IlyaSutskever2014SequenceTS}, \nocite{fei2022mutual,FeiMatchStruICML22,FeiLasuieNIPS22,wu2023nextgpt} abstractive headline generation have achieved the current SoTA performances \cite{song2020attractive,wang-etal-2022-timestep}.


However, it has been demonstrated that headline generation is much different from the general text summarization problem \cite{xu-etal-2019-clickbait,song2020attractive}.
Essentially, summary text requires covering all the key information expressed in the raw document, and meanwhile the resulting summary usually consists of several sentence pieces (e.g., abstract text).
On the contrary, a headline text needs to reflect the core gist of the document with only one sentence or clause, which can be much more condensed than a summary.
As uncovered, both summary and headline texts should follow the \emph{truthfulness} characteristic, while headline generation additionally needs to satisfy the \emph{attractiveness}.
Thus, many efforts strive to enhance these two attributes for improving headline generation. 
The main motivation is to enhance the semantic understanding of the input documents.
\citet{sun-etal-2015-event} propose identifying the key event chain features for more faithful headline generation.
More external features are incorporated to improve the truthfulness of headlines \cite{matsumaru-etal-2020-improving,wang-etal-2022-timestep}.
Recently, \citet{song2020attractive} balance the headline attractiveness and truthfulness via reinforcement learning. 
\citet{Li-etal-2023-Compressed} employ the construction of a multi-document heterogeneous graph to capture the diversity of relationships within documents, enabling the generation of summaries. \nocite{wu-etal-2023-cross2stra,fei2021optimizing,li-etal-2023-diaasq,shi-etal-2022-effective,Wu0LZLTJ22,Wu0RJL21,FeiGraphSynAAAI21,ZhuangFH23,zhao-etal-2023-generating-visual}
In this work, we take into account the lengthiness and hybridity nature of the document, modeling the discourse structure for fully capturing the underlying semantics.

\subsection{Discourse Structure Modeling}

The external structural information has been frequently employed and integrated into text generation tasks so as to model the intrinsic structure of texts.
Document texts are often characterized by lengthy contents, which inevitably lead to long-range dependency issues \cite{xu-etal-2020-discourse}.
Therefore, for headline generation, it is indispensable to model the underlying structure.
Overall, two granularities of structures need to be taken into consideration for modeling the overall documents, including the document-level and the sentence-level structure.

RST is proposed by \citet{mann1988rhetorical}, in which a document can be presented by a hierarchical discourse tree.
RST-style trees advance in expressing the organization of EDUs in documents, where such document-level discourse information can be leveraged for downstream tasks.
\citet{gerani-etal-2014-abstractive} use the discourse structures and the rhetorical relations of documents for a summary generation. 
\citet{xu-etal-2020-discourse} employ the RST trees to capture long-range dependency for summarization.
\citet{adewoyin-etal-2022-rstgen} encode RST discourse structures for the semantics and topic feature learning for text generation. 
This work follows the same practice, making use of the RST tree to represent the document discourse structure.

There is a list of sentence-level structure theories that are frequently considered for text generation.
From the syntactic perspective, such as the dependency trees \cite{BanerjeeMS15, JinW020} and the phrasal constituent trees \cite{li-etal-2014-improving} are the commonly utilized structural features, which describe how the token pieces are connected and organized into the clauses and sentences.
However, these structures only depict the surface form of word tokens, which fail to capture the underlying semantic structures.
AMR \citep{banarescu2013abstract} offers an option to better represent the abstract semantic meaning of sentences.
\citet{liu-etal-2015-toward} create AMR graphs to help generate the summary. 
\citet{takase-etal-2016-neural} encode the AMR feature with a Tree-LSTM model to improve headline generation. 
\citet{khan2018abstractive} propose a semantic graph-based approach for a multi-document abstractive summary generation. 
Thus, we also consider taking the advantage of the AMR structure.
We construct a unified semantic discourse structure where the document-level global discourse is modeled by the RST structure and the sentence-level AMR structure is responsible for capturing the fine-grained local semantics.

\begin{figure}[!t]
\centering
\includegraphics[width=0.98\columnwidth]{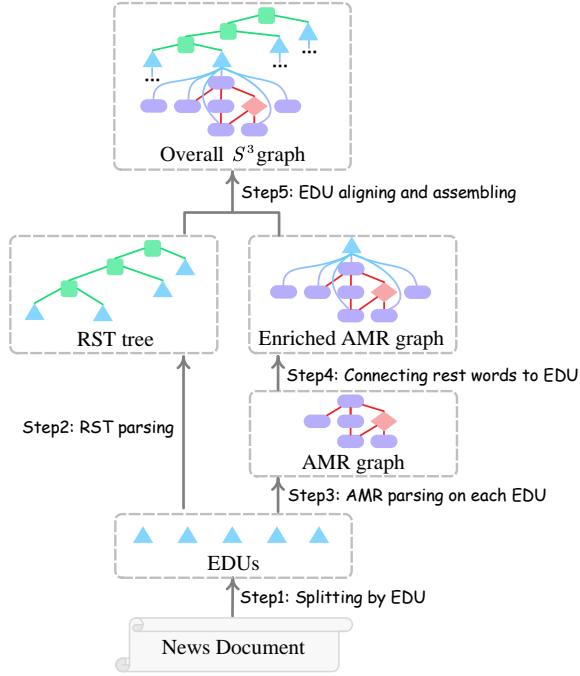}
\vspace{-2mm}
\caption{
The procedure of the S$^3$ graph construction.
}
\label{fig:constrcution}
\end{figure}

\section{Discourse Structure Construction}
\label{sec:doc_graph_construct}

In this section we elaborate on the construction of the S$^3$ graph.
We first give brief definitions of the two meta-structures, RST and AMR structures, based on which the S$^3$ graph is constructed.

\vspace{-1mm}
\subsection{Formulation of Structures}

\vspace{-2mm}
\paratitle{Document-level RST Tree.} 
As shown in Figure \ref{discourse-graph}, a document $D$ can be divided into a set $[e_1, \cdots, e_i, \cdots]$ of smaller elementary discourse units (i.e., EDUs), which can be sentences or clauses.
Then, all EDUs (terminal nodes) are connected and composed into bigger spans (non-leaf nodes e.g., $e_{i:j}$), forming a tree-like structure.
Specifically, given two EDUs, one of them has a specific role relative to the other, which is a pre-defined rhetorical relation, e.g., `\emph{Elaborate}', `\emph{Joint}' and `\emph{Background}'.\footnote{
In \url{https://is.gd/CKArOg} we detail all the rhetorical relation labels.
}
Within the RST tree, each EDU has a `nucleus' or `satellite' identity under the rhetorical relation, indicating the dominance and dependency role of each.

\vspace{-2mm}
\paratitle{Sentence-level AMR Graph.} 
For each elementary discourse
units $e_i$ in $D$, the AMR structure is defined as a directed and labeled graph, where the nodes represent concepts and the edges represent relationships between concepts. 
As illustrated in Figure \ref{discourse-graph}, a node is either a real word in $e_i$ or a dummy (virtual) node, where the latter is added to represent some complex or implicit relationships between other concepts.
Also, not all the individual words in a sentence are included in an original AMR.
Nodes are connected with the `predicate-argument' relation\footnote{
\url{https://is.gd/CKArOg} lists the commonly occurring AMR relation labels.
}, describing the abstract semantic meaning of sentences.
This way, sentences with the same meaning come with the same AMR, even if they are not identically worded, e.g., 
`\emph{The boy desires the girl to believe him}' and `\emph{The boy has the desire to be believed by the girl}'.

\subsection{Constructing Document Graph}
\label{Constructing Document Graph}

Given a document $D$, we now construct the unified semantic discourse structure S$^3$ graph following the main five steps, as illustrated in Figure \ref{fig:constrcution}.

\paratitle{Step1: Splitting document into EDUs}.
First, we adopt an EDU segmenter \citep{zhang2020syntax} to split $D$ into a set $[e_1, \cdots, e_i, \cdots]$ of EDUs.

\paratitle{Step2: Parsing RST structure}.
For all EDUs, we utilize a SoTA RST parser \cite{zhang-etal-2021-adversarial} to build the RST tree structure, marked as $G^R$=<$V^R, E^R$>, where $V^R$ is the set of RST nodes (i.e., EDUs and text-spans), $E^R$ is the set of relational edges.

\paratitle{Step3: Parsing AMR structure}.
At the same time as the step2, we use an off-the-shelf AMR parser \citep{bai-etal-2022-graph} to obtain AMR graph $G^A_i$=<$V^A, E^A$> for each EDU $e_i$.
$V^A$ is the node set (including partial words and virtual concepts), and $E^A$ is the edge set.

\paratitle{Step4: Connecting rest words to EDU}.
We now connect all the word nodes in AMR graph $G^A_i$ to the corresponding EDU node $e_i$.
Such edges are named as `RST-AMR relation', which are also attached with a special label `\emph{RST-AMR}'.
Some words in EDU text are not included in $G^A_i$, which potentially may cause information missing.
Therefore, we also connect the rest words in the EDU text to the EDU node.

\paratitle{Step5: Aligning EDU nodes and assembling graph}.
Finally, we assemble all the AMR graphs $G^A_i$ and the RST tree $G^R$ together by aligning the EDU nodes, yielding the final S$^3$ graph.
We denote the overall graph as $G$=<$V,E$>, where the node sets $V$=$V^R \cup V^A$, and $E$=$E^R \cup E^A$.
Note that the edges in $E$ come with labels.

It is also noteworthy that we categorize the nodes into three hierarchical types from top to bottom:
A-type refers to the Text-span nodes within the EDU part structure;
B-type refers to the EDU nodes (i.e., sentence or clause level);
C-type represents the word-level nodes, including the words and dummy nodes.

\begin{figure*}[!t]
\centering
\includegraphics[width=0.98\textwidth]{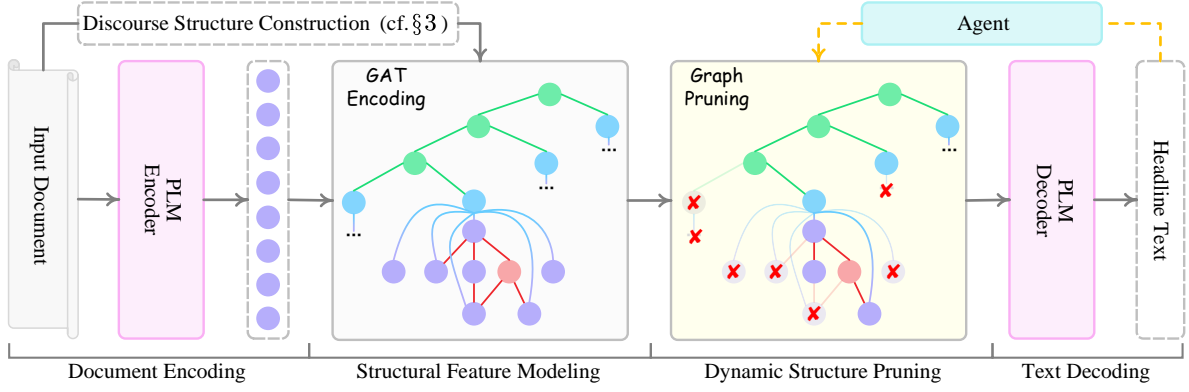}
\vspace{-3mm}
\caption{
Overview of our framework.
A PLM first encodes the input document into representations.
Then, a GAT model encodes the discourse structure S$^3$ of the document for structural feature modeling.
Thereafter, the proposed punning mechanism dynamically filers those less-informative structures within S$^3$ for discourse feature refining.
Finally, the resulting contextual features are integrated into the PLM decoder for headline generation.
}
\vspace{-2mm}
\label{fig:framework}
\end{figure*}

\section{Headline Generation Framework}

Given an input document $D$, our aim is to generate a headline text $y$ for $D$.
As shown in Figure \ref{fig:framework}, our system mainly comprises four modules: document encoding, structural feature modeling, dynamic structure pruning and text decoding.
Following we detail each part.

\vspace{-2mm}
\subsection{Document Encoding}

We first adopt a PLM to represent the overall words of input document texts into contextual representations.
Instead of feeding the whole document texts into PLM, we use a `[SEP]' token to separate each EDU text:
\begin{equation} \nonumber
X = \text{[CLS]}\, \underbrace{w_1\, w_2\,\cdots }_{e_1}\, \text{[SEP]}\, \underbrace{w_j\, w_{j+1}\, \cdots }_{e_i}\, \text{[SEP]} \, \cdots 
\end{equation}
Here $w_i$ is the word token.
`[CLS]' is a special token representing the overall input texts.
Then, PLM encodes the overall text piece:
\begin{equation} \label{PLM}
\mathbf{H} = \{\cdots, \mathbf{h}_{i}, \cdots\}=  \text{PLM}(X) \,, 
\end{equation}
where $\mathbf{h}_{i}$ is the representation of word text $w_i$.

\subsection{Structural Feature Modeling}

In this step, we integrate the document discourse structure features into the context features.
Specifically, we encode the S$^3$ graph $G$=<$V,E$>.
We denote $M$=$\{m_{i,j}\}_{n\times n}$ ($n$ is the total number of nodes) as the adjacent matrix of $G$, with each $m_{i,j}$ corresponding to the edge in $E$.
$m_{i,j}$=1 indicates that there is an edge between node $v_i$ and $v_j$, otherwise $m_{i,j}$=0.
In addition, to enhance the message passing, we construct the reverse edge and self-join link for each node.
The embeddings of word nodes in $G$ are initialized by the contextual representations from PLM from Eq. (\ref{PLM});
for the dummy nodes in $G$, the embeddings are randomly initialized.
For the RST nodes, we create their embeddings by pooling the text spans over their containing words.
Without losing generality, we unifiedly denote all node embeddings as $\bm{r}^v_i$.
For the associated edge labels in $E$, we also maintain their label embeddings $\bm{r}^m_{i,j}$.

Now we employ a $K$-head GAT to model the node in $G$:
\begin{equation}\label{eq:GAT}
\mathbf{u}_i = \sigma (\frac{1}{K} \sum_{k=1}^K  \sum_{j \in \mathcal{N}_i } \beta_{i,j}^k \mathbf{W}^k (\bm{r}^v_i \oplus \bm{r}^v_j \oplus \bm{r}^m_{i,j}) ) \,, \\
\end{equation}
where $\mathbf{W}^k$ is parameter,
$\sigma$($\cdot$) is the sigmoid activation function,
$\oplus$ is element-wise addition,
$\beta_{i j}^k$ is normalized attention weight computed by the $k-$th attention head:
\setlength\abovedisplayskip{2pt}
\setlength\belowdisplayskip{2pt}
\begin{equation}
\beta_{i,j}^k =   \frac{m_{i,j} \, \exp (\sigma(\bm{r}^v_i \oplus \bm{r}^v_j \oplus \bm{r}^m_{i,z}))}{\sum_{z \in \mathcal{N}_i}\,  m_{i,z}\,   \exp (\sigma(\bm{r}^v_i \oplus \bm{r}^v_z \oplus \bm{r}^m_{i,z}))} \,.
\end{equation}

\begin{figure}[!t]
\centering
\includegraphics[width=0.98\columnwidth]{figures/pruning3.pdf}
\vspace{-3mm}
\caption{
Illustration of the pruning mechanism.
}
\label{fig:pruning}
\vspace{-8mm}
\end{figure}


\vspace{-1mm}
\subsection{Dynamic Structure Pruning}

A long document contains a vast volume of information.
Intuitively thinking, there can be only a small part of features within the graph that really contribute to the headline generation, i.e., providing crucial features for indicating the core semantics; while a majority of structures can be seen as superfluous details.
This motivates us to devise a mechanism of structural pruning over the S$^3$ graph, dynamically filtering those noise nodes.
Specifically, given the initial S$^3$ graph with all nodes, we consider the pruning behavior as a binary decision, i.e.,  whether drop the node or not. 
To implement the idea, we formulate the node selection as a Markov decision process, and apply the REINFORCE algorithm \citep{williams1992simple}.

Technically, we build a 3-layer feedforward network (FFN) as the RL agent $\pi$($\bm{s};\theta$) ($\theta$ is the parameter) for generating the pruning decision.
The agent takes a graph node representation as the state feature, i.e., $\bm{s}_i$=$\bm{u}_i$.
and generate an action for node $v_i$.
An action is a pruning probability to $v_i$, which is sampled from a Gaussian
distribution:
\setlength\abovedisplayskip{2pt}
\setlength\belowdisplayskip{2pt}
\begin{equation}
a_i \sim  \pi(\bm{s}_i;\theta)=\mathcal{N}(0,I) \,.
\end{equation}
Nodes at different granularities within the discourse structure serve different extents of roles.
For example, most of the word nodes should be aggressively screened, while pruning the EDU nodes should be cautious.
Intuitively, the higher the level, the more important the node, or vice versa.
Thus, we design a hierarchical pruning manager.
According to the granularities of nodes (A/B/C-type), we design three levels of thresholds hierarchically, i.e., $p^A$, $p^B$ and $p^C$, with $a_i$ larger than the corresponding type of threshold indicating pruning away the node from $G$.
In Figure \ref{fig:pruning} we illustrate the overall pruning mechanism.

It is a tentative pruning process with one action each time, where the goal is to find the best policy $\pi$ to prune $G$ for the best task performance.
We evaluate the pruning effect by observing the quality of the resulting headline.
Specifically, we create two types of RL rewards, the cross-entropy value and the ROUGE score, where the former indicates the confidence of current task prediction $R^c$=$\log p(y|\hat{G}^l)$ ($\hat{G}^l$ represents the $l$-th pruned S$^3$ graph), and the latter indicates the improvement over the original unpruned graph $R^r$=ROUGE($\hat{G}^l$)-ROUGE($G$).
The total reward is written as $R$=$R^c+R^r$.
By iteratively executing the above pruning process $l$ times, we finally obtain a refined and structurally compact discourse structure $\hat{G}^l$.
We encode the structure with Eq. (\ref{eq:GAT}) to obtain the enhanced contextual features $\{\bm{z}_i\}$.
The agent is updated and optimized with the Monte-Carlo policy gradient algorithm:
\setlength\abovedisplayskip{2pt}
\setlength\belowdisplayskip{2pt}
\begin{equation} \label{pruning-pg}
\nabla J(\boldsymbol{\theta})=\mathbb{E}_{s, a \sim \pi}\left[R_\pi\left(s^l, a^l\right) \frac{\nabla \pi\left(a^l \mid s^l, \boldsymbol{\theta}\right)}{\pi\left(a^l \mid s^l, \boldsymbol{\theta}\right)}\right] \,.
\end{equation}

\begin{table*}[!t]
\caption{
Results on CNNDM-DH dataset. 
Scores with $^\dag$ are retrieved from \citet{song2020attractive}. 
}
\label{tab:main1}
\resizebox{0.99\textwidth}{!}{
\begin{tabular}{lccccccccc}
\toprule
\bf Methods & \bf BLEU-1 & \bf BLEU-2 & \bf BLEU-3 & \bf BLEU-4 & \bf ROUGE-1 & \bf ROUGE-2 & \bf ROUGE-L & \bf METEOR & \bf\em Avg.\\
\midrule
\multicolumn{9}{l}{\textbf{$\blacktriangleright$ \emph{W/o using PLM}}}\\
Seq2seq$^\dag$  &-&-&-&- & 13.57 & 3.07 & 11.82 & -& -\\
PGN$^\dag$ &-&-&-&-& 27.80 & 12.06 & 23.30 & -& -\\
FAST-ABS$^\dag$ &-&-&-&-& 34.84 & 15.91 & 30.26 & 17.28& -\\
PORL-HG$^\dag$ &-&-&-&- & 34.23 & 15.35 & 29.48 & 18.07& -\\
\cline{1-10}
\multicolumn{9}{l}{\textbf{$\blacktriangleright$ \emph{Using PLM}}}\\
PORL-HG  &30.84&20.96&14.61&10.53& 37.73 & 17.48 & 32.58 & 20.63 & 23.17 \\
BART  &31.22 & 21.39 & 14.93 & 10.91 & 37.80 & 17.54 & 32.67 & 20.98& 23.43\\
HipoRank & 32.81 & 22.04 & 15.41 & 11.67 & 38.14 & 17.85 & 33.08 & 21.07&24.01 \\
TSE-AC & 34.08 & 23.90 & 16.74 & 12.85 & 38.42 & 18.17 & 33.48 & 20.26& 24.73 \\
\bf Ours & \textbf{38.96} & \textbf{28.07} & \textbf{20.19} & \textbf{15.15} & \textbf{40.83} & \textbf{20.49} & \textbf{36.44} &\textbf{21.34}& \bf 27.68 \\

  \bottomrule
\end{tabular}
}
\vspace{-2mm}
\end{table*}

\vspace{-4mm}

\subsection{Text Decoding}

During the decoding stage, we integrate the above-obtained discourse structure features.
Specifically, we adopt the cross-attention \cite{vaswani2017attention} to fuse the three representations together into the PLM decoder:
\setlength\abovedisplayskip{2pt}
\setlength\belowdisplayskip{2pt}
\begin{equation}
\bm{C} = \text{Softmax} \left( \frac{\bm{Z}^T \cdot  \overline{\bm{Z}}}{\sqrt{d}}  \right) \cdot \bm{O} \,,
\end{equation}
where $\bm{Z}$=$\{\bm{z}_i\}$ is the overall pruned graph representation, $\overline{\bm{Z}}$ is the representation of real word nodes, a subset of $\hat{G}^l$ graph.
And $\bm{O}$ is the decoder representation.
With the final representation $\bm{C}$, PLM decoder autoregressively generates the headline text:
\setlength\abovedisplayskip{2pt}
\setlength\belowdisplayskip{2pt}
\begin{equation}
\hat{y}=\operatorname{argmax}_y \prod_{t=1}^T P\left(y_t \mid C, \hat{y}_{<t}\right) \,.
\end{equation}

\subsection{Training Details}
The main target is to minimize the cross-entropy loss of the headline generation towards the ground-truth texts:
\setlength\abovedisplayskip{2pt}
\setlength\belowdisplayskip{2pt}
\begin{equation}\label{CE-loss}
\mathcal{L}_{\text{CE}} = \sum \log p(y|D) \,.
\end{equation}
Meanwhile, we also need to optimize the pruning module via Eq. (\ref{pruning-pg}).
In practice, a warm-start training strategy is adopted, by first training the task cross-entropy loss, and once it tends to converge, we then join the pruning training together, until the whole system reaches its plateau.

\section{Experimental Settings}

\begin{table*}[!t]
  \caption{Results on DM-DHC dataset, where all methods use the same BART PLM.}
  \label{tab:main2}
\resizebox{0.99\textwidth}{!}{
  \begin{tabular}{lccccccccc}
    \toprule
    \bf Methods & \bf BLEU-1 & \bf BLEU-2 & \bf BLEU-3 & \bf BLEU-4 & \bf ROUGE-1 & \bf ROUGE-2 & \bf ROUGE-L & \bf METEOR & \bf\em Avg.\\

    \midrule
    PORL-HG  & 30.43 & 20.79 & 14.21 & 10.35 & 37.58 & 17.42 &32.41 & 20.42 & 22.95\\
    BART & 30.78 & 20.92 & 14.53 & 10.49 & 37.67 & 17.46 & 32.53 & 20.74 & 23.14\\
    HipoRank & 32.36 & 21.57 & 15.12 & 11.29 & 38.06 & 17.72 & 32.91 & 21.03&23.76 \\
    TSE-AC & 33.46 & 23.63 & 16.28 & 12.38 & 38.23 & 18.06 & 33.31 & 20.19  & 24.44\\
    \bf Ours & \textbf{37.85} & \textbf{27.12} & \textbf{18.76} & \textbf{14.52} & \textbf{40.54} & \textbf{19.98} & \textbf{36.14} & \textbf{21.29} & \textbf{27.03}\\
  \bottomrule
\end{tabular}
}
\vspace{-2mm}
\end{table*}

\subsection{Dataset and Implementation}
Our experiments are based on two headline generation datasets, including the CNNDM-DH \citep{Nallapati2016abstractive} and DM-DHC \citep{song2020attractive}, where the documents derive from CNN or Daily Mail news articles. 
The instances are split into train/develop/test sets as 281,208/12,727/10,577 in CNNDM-DH and 138,787/11,862/10,130 in DM-DHC, respectively.

For each document, we obtain the RST trees via a pre-trained off-the-shelf RST parser \citep{zhang-etal-2021-adversarial}, where we preserve the tree topology, EDUs segmentation, node type, and node relations for each document. 
The AMR graphs are acquired by a SoTA AMR parser \citep{bai-etal-2022-graph}.
In addition, to match AMR words with the actual word in the raw document, we apply NLTK\footnote{
\url{https://www.nltk.org}
} 
and the Spacy\footnote{
\url{https://spacy.io}
} 
toolkit for preprocessing.

Our system takes the BART-base\footnote{
\url{https://huggingface.co/docs/transformers/model_doc/bart}
} as the backbone PLM.
The RL agent has 3 layers with 300-d hidden size.
The hierarchical thresholds for pruning are set as: $p^A$=0.85, $p^B$=0.60 and $p^C$=0.40 and these values are shown to give the best results in our preliminary experiments.
Adam \citep{kingma2014adam} optimizer is used for training,
where the learning rate is 5e-6 for PLM backbone and agent, but 5e-4 for GAT.
The dropout ratio is set to 0.1 universally. 
We use beam search with size 2 to generate the headlines.

\vspace{-3mm}
\subsection{Baseline and Evaluations}

We mainly compare with the following existing strong-performing headline generation systems.
\begin{compactitem}
    
   \item  \textbf{PGN} \citep{see-etal-2017-get}, which is a pointer network-based text generation system.
   
   \item  \textbf{FAST-ABS} \citep{chen-bansal-2018-fast} generates headline texts by extracting salient sentences and rewriting them via the RL method. 

    \item  \textbf{PORL-HG} \citep{song2020attractive} based on the FAST-ABS model, further adding the news article comment information.

    \item  \textbf{HipoRank} \cite{dong-etal-2021-discourse} exploits the hierarchical graph representations for the source document for title generation.

    \item  \textbf{TSE-AC} \citep{wang-etal-2022-timestep} enhances the headline generation by utilizing the timestep-aware sentence embedding, and relocating the critical words during decoding.

\end{compactitem}

For fair comparisons, we also re-implement the strong-performing baselines by integrating the same BART-base PLM.

Following \citet{song2020attractive}, we adopt three types of metrics for performance evaluation, including BLEU(-1/2/3/4) \cite{papineni2002bleu},
ROUGE(-1/2/L) \citep{lin2004looking}
and METEOR \citep{denkowski2014meteor}.
Note that scores from our implementation are the average over five runs with the latest given checkpoints.

\section{Results and Discussion}

\subsection{Overall Comparisons}

\paratitle{Results on CNNDM-DH data.} 
Table \ref{tab:main1} presents the overall comparisons of CNNDM-DH data.
As can be found, with additional feature modeling, the performances are enhanced (i.e., PORL-HG \& FAST-ABS vs. PGN).
In particular, the use of PLM significantly improves the results of title generation. 
Under the fair comparison of using the same PLM, we can find that our overall system evidently beats all the baseline models on BLEU METEOR and ROUGE metrics consistently.
Specifically, we outperform the best baseline (TSE-AC) with a 2.95(=27.68-24.73) average score of all metrics, and especially improve the BLEU-1 and BLEU-2 by 4.88(=38.96-34.08) and 4.17(=28.07-23.90) score, respectively.
We notice that given SoTA baselines (e.g., PORL-HG, HipoRank and TSE-AC) integrate the features to help spot the key feature words within documents, our models still yield better results.
Therefore, we speculate that focusing on the core semantic information of documents is better than focusing on specific critical words in generating high-quality headlines.  
The above overall comparisons directly demonstrate the efficacy of our system.

\begin{table}[!t]
  \caption{Human evaluation results on the quality of generated headlines.
  Our system vs. TSE-AC (PLM) \citep{wang-etal-2022-timestep}.
  For the last item, disabling both the pruning mechanism and the S$^3$ graph is equivalent to the vanilla BART.
  }
\label{tab:Human}
 \setlength{\tabcolsep}{0.7mm}
\resizebox{0.98\columnwidth}{!}{
\begin{tabular}{lccc}
\toprule
\bf Methods & \bf Relevance($\uparrow$) & \bf Attraction($\uparrow$) & \bf Fluency($\uparrow$)\\
\midrule
Groundtruth & \bf 4.39 & 3.82 & \bf 4.63\\
\cdashline{1-4}
TSE-AC (PLM) & 4.07 & 3.69  &  4.41\\
Ours & 4.26 & \bf 3.93 & 4.60 \\
\quad w/o pruning & 4.18 & 3.77 & 4.52\\
\qquad w/o S$^3$ graph & 3.93 & 3.44 & 4.32 \\
\bottomrule
\end{tabular}
}
\vspace{-4mm}
\end{table}

\paratitle{Results on DM-DHC Data.} 
Table \ref{tab:main2} further shows the performances on DM-DHC dataset.
Overall, we observe a similar trend with that in Table \ref{tab:main1}, where our model achieves the best performance against all comparing systems.
In particular, ours outperforms the TSE-AC by a 2.59(=27.03-24.44) average score on all metrics.
We can give the credit to our method of capturing the core document semantics by integrating the proposed discourse structure features.

\paratitle{Human Evaluation Results.}
In addition to the automatic metrics, here we also provide human evaluation, directly examining the detailed quality of the generated headline texts.
We design the Likert 5-scale in terms of the relevance, attraction and fluency of texts,
Then, we ask three English native speakers to assess the generation quality by different models.
We randomly select 100 testing samples from CNNDM-DH.
Table \ref{tab:Human} shows the averaged scores on three aspects.
We first see that the human-annotated (ground-truth) headlines are with the highest semantic relevance and great language fluency, while lacking attraction.
On the contrary, our system helps yield high-quality headlines in terms of all three aspects, where notedly, we even achieve better attraction.
Moreover, the best baseline (TSE-AC) exhibits inferior generation quality, compared with ours.
Also, without using the proposed S$^3$ graph features (i.e., vanilla BART), all three aspects deteriorate significantly, while removing the pruning mechanism slightly hurts the overall quality.

\subsection{System Ablation}

We consider quantifying the specific contribution of each part of our method.
Thus we present the ablation study over the S$^3$ graph and the neural framework, as shown in table \ref{tab:ablation}.
First of all, by disabling the dynamic pruning module from our system, we can witness clear drops, i.e., around 1 average point in all metrics.
If disengaging the hierarchical pruning strategy, the final performance is also affected (not much as the whole pruning ablation), which proves the necessity of designing a hierarchical pruning mechanism over S$^3$ graph.

Further, we examine the effectiveness of the proposed S$^3$ graph. 
To avoid the influence of the pruning module, we run the experiment based on the variant model without equipping the pruning mechanism.
First, either removing away the sentence-level AMR graphs or the document-level RST trees from the unified S$^3$ graph will hurt the efficacy.
In particular, RST shows more significant influences than the AMR structures.
This can be reasonable because although the AMR effectively represents the fine-grained semantic meaning of texts, RST trees that describe the overall discourse relations within the whole documents can play a larger role in document semantic understanding.
However, canceling the whole S$^3$ graph features for headline generation, we witness the biggest performance drops, i.e., with -2.79(=21.52-24.31) on average, and especially with -4.24(=10.91-15.15) BLEU-4.

\begin{table}[!t]
\caption{The ablation results on CNNDM-DH dataset.
`HPM' denotes the hierarchical pruning manager.
`w/o AMR' means constructing alternative structures that only use the RST trees to represent the document EDUs.
`w/o RST' means representing documents with AMR graphs for all sentences.
}
\label{tab:ablation}
\resizebox{0.98\columnwidth}{!}{
\begin{tabular}{lcccc}
\toprule
\bf Methods & \bf BLEU-4 & \bf ROUGE-L & \bf METEOR &\bf\em Avg.\\
\midrule
\bf Ours (Full) & \textbf{15.15} & \textbf{36.44} & \textbf{21.34} & \textbf{24.31} \\
\cdashline{1-5}
\quad w/o Pruning& 14.09 & 35.13 & 21.25 & 23.49 \\
\qquad w/o HPM & 14.32  & 35.91  & 21.27  & 23.83   \\
\cdashline{1-5}
\qquad w/o AMR & 12.46 & 33.37 & 21.18 & 22.34\\
\qquad w/o RST & 11.23 & 32.95 & 21.06 & 21.75 \\
\qquad w/o S$^3$ graph & 10.91 & 32.67 & 20.98 & 21.52\\
\bottomrule
\end{tabular}
}
\vspace{-3mm}
\end{table}

\subsection{Further Analyses}

Above we demonstrate that our model achieves the SoTA performance on the headline generation.
In this part, we take one step further, investigating how our proposed methods advance the task.

\begin{figure}[!t]
\centering
\includegraphics[width=0.98\columnwidth]{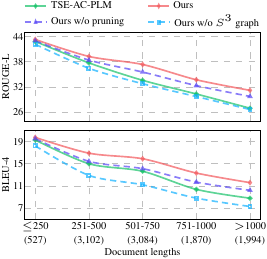}
\caption{
Generation performance under various lengths of input documents.
In the brackets are the document numbers.
}
\label{fig:lengths}
\end{figure}

\paratitle{Impact of Documents Length.}
Intuitively, longer documents cover more background information that is less useful to the final title generation, and correspondingly burdens the model performances.
Thus we study the impact of document length.
In Figure \ref{fig:lengths} we show the analysis results.
As seen, there is a consistent trend that the overall performance of each model decreases gradually when the documents have longer contents.
For short-text-length documents containing less than 250 words, the performance gap can be small between our model and the TSE-AL system.
However, for the longer document, especially the super lengthy ones (i.e., $>$750), the advantages of our model become clearer.
However, without integrating the discourse S$^{3}$ graph features, the ability of our model on adapting to the document length growth deteriorates significantly.
Overall, this reveals that the discourse structure information effectively helps in handling the long documents, which is consistent with prior relevant findings \citep{xu-etal-2020-discourse}
Besides, compared with the discourse structure features, the pruning mechanism shows lower impacts.

\begin{figure}[!t]
\centering
\includegraphics[width=0.98\columnwidth]{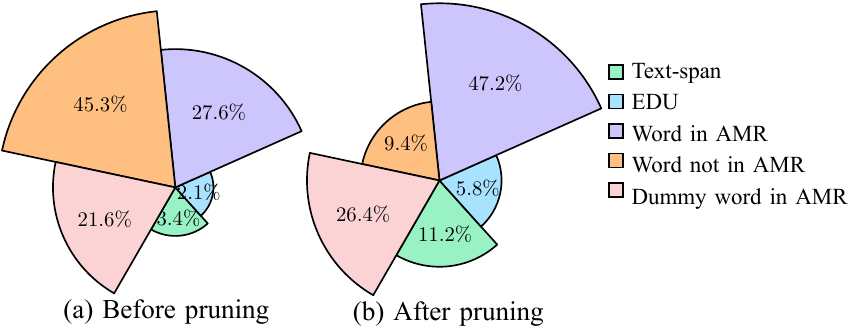}
\caption{
The percentage of different types of nodes in S$^3$ before (a) and after (b) pruning.
}
\vspace{-3mm}
\label{fig:node_change}
\end{figure}

\begin{figure*}[!t]
\centering
\includegraphics[width=0.98\textwidth]{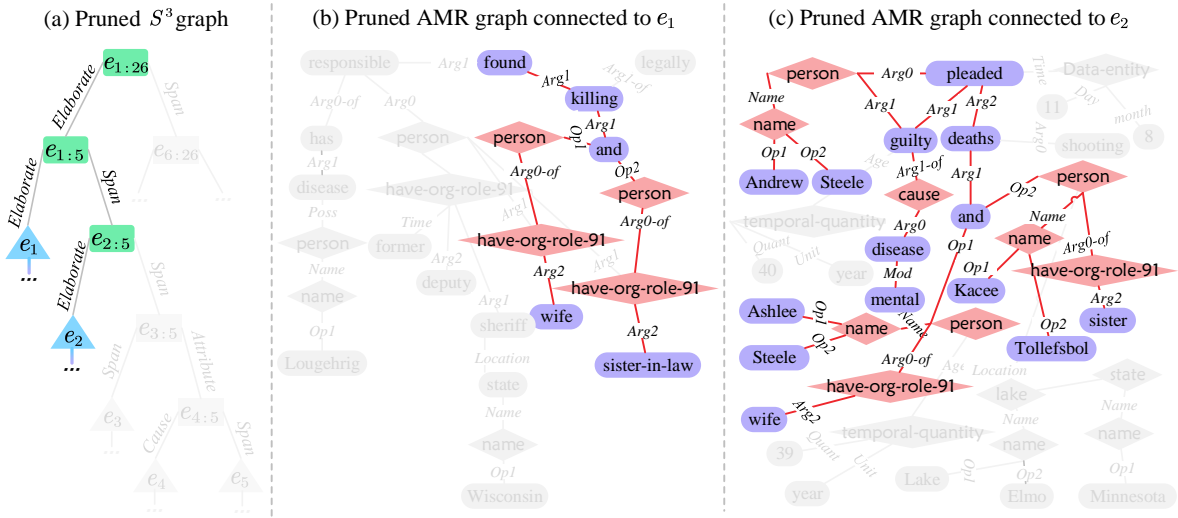}
\caption{
Quantitative result of our framework.
The document instance is randomly selected from the CNNDM-DH test set.
The colored nodes are retained after pruning, while the pruned structures are in grey.
}
\label{fig:case}
\vspace{-1mm}
\end{figure*}

\paratitle{Roles of Different Structure Types in S$^3$.}
In our S$^3$ graph, combing both the RST tree and AMR graphs, we have multiple types of nodes, while each of the structure types can serve varied roles within the whole structure.
Here we explore the distinction of the structure roles.
Specifically, we make statistics of the following nodes within the graph: text-span node, EDU node, AMR real-word node, the real word not belonging to AMR\footnote{
As stated in step 4 at $\S$\ref{Constructing Document Graph}, some word tokens are not included in AMR graphs, which are also linked to EDU nodes within the S$^3$ graph.
}, AMR dummy node.
Since the resulting S$^3$ graph after pruning gives the best effect, we consider comparing the constituent proportions of nodes before and after the pruning, respectively.
Figure \ref{fig:node_change} displays the trends.
We can see that before pruning, the words not belonging to AMR take the major proportion.
However, most of these types of nodes can be the background noise carrying less information to represent the core document semantics, thus contributing less to the final headline generation. 
Correspondingly, after the pruning process, this node type is dramatically filtered out, from 45.3\% to 9.4\%.
On the contrary, the proportion of the type of `words in AMR' increases significantly.
These words together with the dummy concepts in AMR graphs are the essential clues words representing the critical abstract meanings, which are beneficial to headline generation.
Besides, at the document level, the weights of the text-span and EDU nodes are highlighted.
These coarse-grained structures (i.e., sentence and clause levels) play the role of discourse planning within the global document, which are important for the overall document semantic understanding.

\paratitle{Impact of Preprocessing Errors.}
To assess the impact of preprocessing errors introduced by the RST parser and AMR parser on the performance of our model, we conducted an empirical analysis utilizing the comparatively suboptimal RST parser and AMR parser. For RST parsing, we employed RST parser proposed by \citet{braud2016multi}, which exhibited a performance degradation of more than 15\% compared to SOTA RST parser \cite{zhang-etal-2021-adversarial} in terms of nuclearity determination, rhetorical relation classification, and full discourse parsing. For AMR parsing, we utilized AMR parser proposed by \citet{zhou2021amr}, which demonstrated a decrease of over 9\% in Smatch score compared to the Bai model we employed. Table \ref{tab:preprocess error} we show the comparison results. We can observe that the utilization of the less effective RST parser and AMR parser leads to a decline in model performance, indicating that certain errors occurring during the RST parsing and AMR parsing processes indeed contribute to a degradation in model effectiveness. However, the decrease in model performance is not excessively severe, suggesting that the pruning process can to some extent counteract the impact caused by parsing errors.

\begin{table}[!t]
\caption{Comparison of results on CNNDM-DH dataset from the inferior parser.
}
\label{tab:preprocess error}
\resizebox{0.98\columnwidth}{!}{
\begin{tabular}{lcccc}
\toprule
\bf Methods & \bf BLEU-4 & \bf ROUGE-L & \bf METEOR &\bf\em Avg.\\
\midrule
\bf Ours & \textbf{15.15} & \textbf{36.44} & \textbf{21.34} & \textbf{24.31} \\
\cdashline{1-5}
\quad \citet{braud2016multi}& 14.23 & 35.31 & 21.29 & 23.61 \\
\quad \citet{zhou2021amr} & 14.89  & 36.08  & 21.32  & 24.10   \\
\bottomrule
\vspace{-6mm}
\end{tabular}
}
\end{table}

\paratitle{Constructing Homogeneous Graph.}
To investigate the influence of node information on model performance, we opted to substitute the heterogeneous graph network within the Graph Attention Network (GAT) with a homogeneous graph network. This alteration was motivated by the necessity to incorporate node information in subsequent layer pruning procedures, necessitating the exclusive application of the homogeneous graph representation method during the GAT phase. Table \ref{tab:homogeneous graph} we show the results. We observe that the adoption of homogenous graphs, as opposed to the previously employed heterogeneous graphs, exhibits various performance declines. This indicates that accounting for the dissimilarities between different nodes facilitates the model's ability to capture the diverse relationships present in the documents. These findings align with the research conducted by \citet{Li-etal-2023-Compressed}., further substantiating the importance of considering node heterogeneity in modeling endeavors.

\begin{table}[!t]
\caption{Comparison with constructing homogeneous graph on CNNDM-DH dataset.
}
\label{tab:homogeneous graph}
\resizebox{0.98\columnwidth}{!}{
\begin{tabular}{lcccc}
\toprule
\bf Methods & \bf BLEU-4 & \bf ROUGE-L & \bf METEOR &\bf\em Avg.\\
\midrule
\bf Heterogeneous Graph & \textbf{15.15} & \textbf{36.44} & \textbf{21.34} & \textbf{24.31} \\
Homogeneous Graph & 14.16 & 35.27 & 21.31 & 23.58 \\
\bottomrule
\vspace{-6mm}
\end{tabular}
}
\end{table}

\paratitle{Comparison with Zero-Shot Large Language Models.}
Currently, numerous open-source large language models \cite{yang2023baichuan, touvron2023llama, bai2023qwen} have demonstrated excellent performance across various general domains. In this study, we aim to compare our model, trained with document structural information, with these zero-shot models, in order to demonstrate the value of utilizing such small models amidst the prevalence of large language models. We conducted experiments using the zero-shot models Baichuan-7B\cite{yang2023baichuan}, Llama2-13b\cite{touvron2023llama}, and Qwen-70b\cite{bai2023qwen}. Table \ref{tab:large language model} we present the results. We can observe that, in relative terms, larger language models possessing a greater number of parameters exhibit superior performance. Nevertheless, even the most proficient model, Qwen-70b, fails to surpass the performance of our model trained with document information. This observation validates the value of constructing document graphs using document information, wherein the utilization of existing document information enables small models to achieve superior performance with reduced computational resources.

\begin{table}[!t]
\caption{Comparison with Zero-Shot Large Language Models on CNNDM-DH dataset.
}
\label{tab:large language model}
\resizebox{0.98\columnwidth}{!}{
\begin{tabular}{lcccc}
\toprule
\bf Methods & \bf BLEU-4 & \bf ROUGE-L & \bf METEOR &\bf\em Avg.\\
\midrule
\bf Ours & \textbf{15.15} & \textbf{36.44} & \textbf{21.34} & \textbf{24.31} \\
\cdashline{1-5}
\quad Baichuan-7b & 5.42 & 23.18 & 16.15 & 14.92 \\
\quad Llama2-13b & 7.00 & 26.41  & 16.98  & 16.80   \\
\quad Qwen-70b & 9.42 & 30.80  & 21.33  & 20.52   \\
\bottomrule
\end{tabular}
}
\end{table}

\paratitle{Case Study.}
Finally, to gain a direct understanding of how our overall method advances the headline generation, we conduct a case study through empirical analysis.
In Figure \ref{fig:case} we display a news document, where our model gives a headline prediction.
Meanwhile, we illustrate the corresponding S$^3$ structure, as well as the pruning behavior over the graph.
First of all, by comparing with the gold headline text, we can notice that our system's prediction is more concise.
Also by reading through the input news article, it is clear that our generated headline has good faithfulness.
On the other hand, the resulting highlighted unified discourse graph after pruning shows a compact yet informative sub-structure.
As seen, at the document level, only small pieces of the text-span nodes and the two associated EDU nodes that are organized with the `\emph{Elaborate}' rhetorical relations are retained, where the corresponding sentences and clauses of raw texts indeed include the key information within the document.
Looking into the two EDU nodes ($e_1, e_2$), the AMR graphs are also refined, with the key parts of the local abstract meaning structures that closely relate to the final headline text are kept.
For example, the resulting AMR sub-graph at $e_1$ depicts the factual event of `\emph{killing wife and sister-in-law}', which is the key content of the headline.
And the rest AMR sub-graph at $e_2$ further offers the key clues of `\emph{Andrew Steele}' and `\emph{not guilty}' to the headline.
This well explains how the document-level and sentence-level structures collaborate together to learn the core semantics of the document.
Overall, the system takes advantage of the unified semantic discourse structure features, and further with the help of dynamic pruning generates high-quality headlines.

\vspace{-1mm}
\section{Conclusion}

In this work, we propose constructing a type of unified semantic discourse structure to represent the core document semantics for high-quality headline generation.
We build the S$^3$ graphs by assembling the document-level RST trees and the sentence-level AMR graphs.
We then develop a headline generation framework, in which we integrate the S$^3$ graph features.
We further devise a hierarchical structure pruning mechanism to filter those redundant and task-irrelevant nodes within the S$^3$ graph.
On two headline generation datasets our method achieve the best performances over all the baselines on both the automatic and human evaluations.
Ablation studies quantify the contribution of each design of our methods.
Further analyses indicate that the discourse structure information especially helps in handling long documents.
A quantitative case study reveals how the S$^3$ graph aids yield high-quality headline texts.

\bibliography{tacl2021}
\bibliographystyle{acl_natbib}








  
\end{CJK}
\end{document}